# Automatic Extraction of Ranked SNP-Phenotype Associations from Literature through Detecting Neural Candidates, Negation and Modality Markers

B. Bokharaeian[1, 2] * and A. Diaz[1]
[1] Facultad de Informática, Universidad Complutense de Madrid, Calle del Prof. José G! Santesmases, Madrid, Spain
[2] Amol University of Special Modern Technologies, Mazandaran, Iran

**Abstract**
Genome-wide association (GWA) constitutes a prominent portion of studies which have been conducted on personalized medicine and pharmacogenomics. Recently, very few methods have been developed for extracting mutation-diseases associations. However, there is no available method for extracting the association of SNP-phenotype from text which considers degree of confidence in associations. In this study, first a relation extraction method relying on linguistic-based negation detection and neutral candidates is proposed. The experiments show that negation cues and scope as well as detecting neutral candidates can be employed for implementing a superior relation extraction method which outperforms the kernel-based counterparts due to a uniform innate polarity of sentences and small number of complex sentences in the corpus. Moreover, a modality based approach is proposed to estimate the confidence level of the extracted association which can be used to assess the reliability of the reported association.

**Keywords:** SNP, Phenotype, Biomedical Relation Extraction, Negation Detection.

## 1.INTRODUCTION

A **single-nucleotide polymorphism** (**SNP**) is a single base mutation that happens in DNA-level [1]. Variations in the DNA sequences can affect how humans develop diseases and respond to pathogens, chemicals, drugs, and other agents. The first successful GWA study dates back to 2005 when Klein and his colleagues carried out the first successful GWAS on patients with age-related macular degeneration. It was the start of a worldwide trend which results in finding thousands SNP associations. Fig 1 shows the increasing numbers of papers that have been published in this field from the year 2001 to 2014 obtained from a PubMed search engine for the query 'Single Nucleotide Polymorphisms' (performed in November 2015). SNPs are also important for personalized medicine.

Recently, few methods have been developed recently for extracting mutation and disease associations from text such as [2] and [3]. However, there is no available method for extracting the association of SNP-phenotype from text which consider the neutral candidates and the level of confidence of associations.

A **phenotype** is the organism's recognizable characteristics or traits, such as its development, biochemical or physiological properties, behavior, and products of behavior [4]. An SNP can be "associated" with the phenotype when a specific type of variant (one allele) is frequent within samples obtained from subjects. The degree in which phenotype is determined by genotype is referred as "phenotypic plasticity" [5].

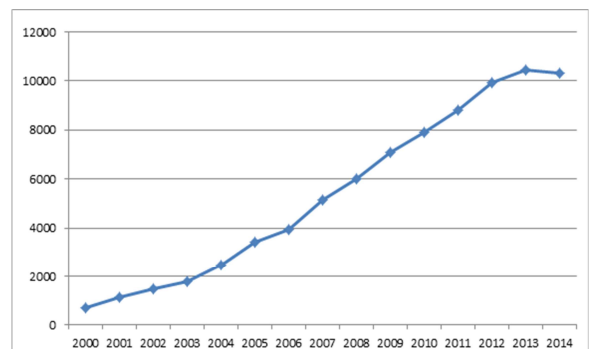

**Figure 1.** Number of 'Single Nucleotide Polymorphisms' publications from 2000 to 2014 in PubMed.

The amount of influence that environmental factors have on a person's ultimate phenotype is a matter of serious scientific debate.

On the other hand, one of the essential tasks in biomedical text mining is to identify negations which is the more important feature used in our approach. Linguists define **negation** as a morphosyntactic operation [6]. Through this operation a lexical item either denies or inverts the meaning of another item or construction. The importance of negation in biomedical text mining is revealed when we consider the fact that negation is very common in those texts leading to lack of precision in automatic information retrieval systems [7]. For example in the sentence below, there is not any association between "*APOE polymorphisms*" and "*serum HDL-C*"; however, if negation is neglected a wrong association might be identified:

- There were <{ no} associations between *APOE polymorphisms* and *serum HDL-C*, APO-CIII and triglycerides>

Linguistic **modality** is another linguistically-driven phenomenon going to be applied in this research. In general, modals are special words stating modality, which expresses the internal attitudes and beliefs of the announcer such as facility, probability, inevitability, commitment, permissibility, capability, wish, and contingency [8]. In current study, we aim to use modals based on linguistic- and speculation analyses for determination of the confidence and strength of the stated SNP-phenotype associations in the corpus.

On the other hand, despite **distinguished** association candidates which include remarks made by the author, a **neutral candidate** does not contain any remarks [10]. In Fig 2, relation status between "anorexia nervosa" and "rs4680" is neutral since the author has not mentioned their association. In other words, any conclusion about the association between these two entities is not possible with this sentence. McDonald *et al.* are one of the very few groups of researchers, who have investigated the neutral candidates in relation extraction task [9]. More information about the neutral relation candidates their important role in the biomedical domain can be found at the other work of the authors [10].

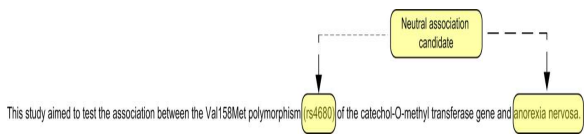

**Figure. 2. A sample of neutral association candidate with used entities specified with circle**

In addition to the pervious subjects, **innate polarity** of the sentences about a relation is an important factor that must be taken into account. However, to the best of our knowledge, no research has been conducted on the effect of innate polarity of the sentences on a relation extraction task.

However, innate polarity of a sentence speaking about a relation expresses whether the assumed relation candidate in the sentence without negation cue and scope exists or not. For instance, the first sample below gives a positive innate polarity on SNP-Phenotype [22], while the second sample provides a negative one.

- The nicotinic acetylcholine receptor polymorphism (*rs1051730*) on chromosome 15q25 is associated with major ***tobacco-related diseases*** in the general population with additional increased risk of COPD as well as lung cancer.
- We investigated the causal relationship between smoking and symptoms of anxiety and ***depression*** in the Norwegian HUNT study using the ***rs1051730*** single nucleotide polymorphism (SNP) variant located in the nicotine acetylcholine receptor gene cluster on chromosome 15 as an instrumental variable for smoking phenotypes.

In this study, we suggest a text-mining approach which extracts association between SNP and phenotypic Phenotypes. The rest of this paper is organized as follows. Section 2 introduces some related research works. The proposed method is explicated in section 3. Afterwards, section 4 presents results and statistical analysis. Finally, section 5 concludes the paper while providing suggestions for further research.

**2. RELATED WORKS**
Besides classical relation extraction tasks in the BioNLP domain such as protein-protein and gen-disease tasks, some new methods and corpora been developed for extracting mutation/polymorphism and disease associations. DiMex [3] is a rule-based unsupervised mutation-disease association extraction that works on the abstract level. The PKDE4J [2] is a supervised method that employs a rich set of rules to detect the used features. Another related miner system has been developed by [15] that gather heterogeneous data from a variety of literature sources in order to draw new inferences about the target protein families.

Moreover, one of the few researches that took **negation** into account in the relation extraction task was [16]. In the method, SVM classifier was fed using a list of features such as nearest verb to candidate entity in the parse tree and some negation cues. Pyysalo *et al.* [18] have conducted a survey wherein negation and **uncertainty** issues were taken into account. They stated that among those corpora **BioInfer** has negative annotation. Numerous studies have been conducted on modality and speculation of identification in NLP [19], but only a few of this research have been employed for classifying speculative language in the bioscience texts. In biomedical study: the vocabularies could be involved in theories, experimental results, hedges, and speculations. Though some studies have been performed within the linguistics community on the use of hedging in scientific text like [20], there is little direct relevance for categorizing task using the perspective of NLP/ML.

## 2. METHOD

The proposed association extraction method relies on detecting linguistic-based negation and neutral candidates which are introduced in this section. The basic components of the algorithm can be seen in the flowchart in Fig. 3.

In this section, the process of detecting SNP-phenotypes associations is explained. It is worth mentioning that we have used the SNPPhenA corpus during the research which has been introduced previously [20]. The corpus is available for public use[1].

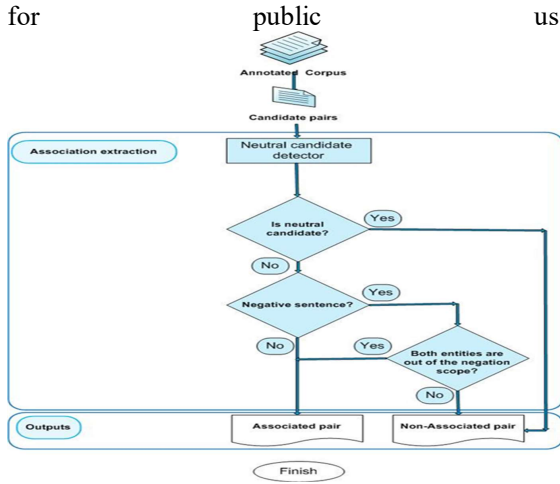

**Figure. 3.** Flowchart of the different steps of the snp-phenotype association extraction proposed algorithm

### 3.1. Verifying the Criteria of the corpus

To examine whether the proposed negation based method is applicable to the corpus or not some metrics must be analyzed which are known as verification criteria (See Fig. 4):

- **Complexity of the sentences:** As mentioned in previous sections, complex sentences form a major source of inaccuracy. They reduce the performance of the algorithm in two ways. Firstly, they decrease the performance of the automatic negation detection algorithm; and secondly, dependent clauses can change the meaning of main clause as mentioned earlier for concessive clauses. Additionally, the number of prominent clause connectors and average number of tokens can be utilized to measure complexity of a sentence.

- **Uniform innate polarity of the sentences regarding to SNP-Phenotype association:** Innate polarity is an important factor in identifying relations from the text. Therefore, the produced corpus is analysed to derive the number of positive and negative innate polarity samples. For an estimation of the ratio of innate positive and negative polarities in the corpus, candidate sentences without negation cue were identified. Selected candidates that express no association between discussed SNP and Phenotype were classified as negatives and the other were identified as positive.

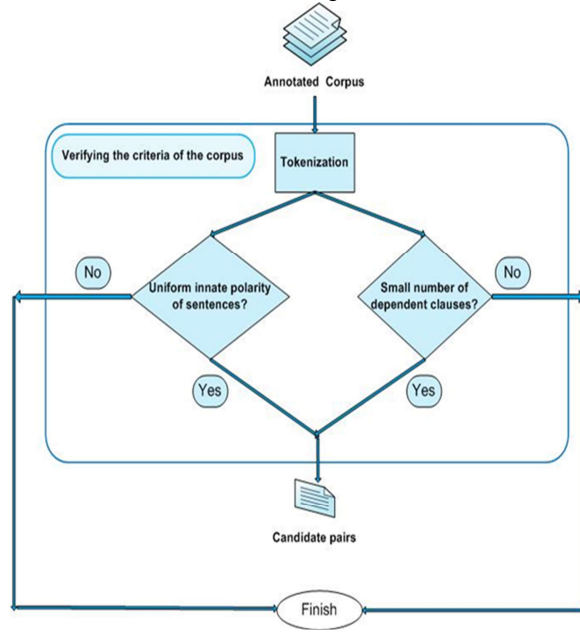

**Figure. 4.** Verifying the criteria of the corpus

### 3.2. Proposed association extraction approach

For developing the proposed approach, six Boolean features were extracted from negation cues and scope which have been used. Additionally, to determine possible effects of negation on SNP-Phenotypes relation, **neutral examples** have been identified in the corpus. Negation inverts status of positive or negative relations candidates which are in the negation scope while leaving neutral ones unchanged. As a result, the ratio of neutral candidates to positive or negative ones is of a great significance.

### 3.2.1. Neutral candidate detector

For automatic detection of neutral candidates we have implemented a neutral candidate detector,

As initial experiments shows detecting the neutral candidates are very important in the negation based method. Consequently a neutral candidate detection system has been carried out. The proposed method worked with a **global context method** kernel method, the prepared corpus has been used for training and predicting the neutral candidates.

---

[1] https://www.dropbox.com/s/2h25suabhrm82ma/SNPPhenA_XML.zip?dl=0

However, in case of neutral candidates, negation does not change the status of the association and it will remain false. Because of few numbers of neutral candidates in the produced corpus, considering the neutral candidates as negatives still leads to superior performance as will be seen in next section (Table 10). As a result, if the status of the existence of neutral candidate was defined as

- "IsNeutralCand" A Boolean feature which is set as true when association candidate predicted as neutral, while other situation is false.

### 3.2.2 Negation based association extraction method

As for relation extraction, it must be noticed that negation does not necessarily change the status of a relation between entities. As a matter of fact, the effect of negation on association depends on several factors among which position of entities relative to the negation scope and cue can be directly extracted from extended corpus. For example, consider the following sentence:

- Moreover, the *rs1051730* variant may *not* merely operate as a marker for *dependence or heaviness of smoking*.

"Dependence or heaviness of smoking" is a phenotypes name inside the negation scope, so as their association relation between SNP (rs1051730) and the phenotype name is inverted by the negation. There are 6 different possibilities based on position of SNP and phenotype names relative to the negation scope which are used as 6 features:

- BothInsNegSc: A Boolean feature which is set as true when both SNP and phenotype names are inside the negation scope, while other situations are false.
- OneLeftOneInsNegSc: A Boolean feature which is set as true when one SNP or phenotype name is on the left side (out) of the negation scope and the other one is inside the negation scope, while other situations are false.
- OneRightOneInsNegSc: A Boolean feature which is set as true when SNP or phenotype name is on the right side (out) of the negation scope and the other one is inside the negation scope, while other situations are false.
- Three other Boolean features related to other possibilities.

As table 1 demonstrates and also we have mentioned earlier, almost all of sentences have positive polarity; hence negation can change the relation status from True to False. Consequently, as it is indicated in Fig. 3 if the studied candidate is not a neutral, and one of these three Boolean features (BothinsideNegSc, OneLeftOneInsideNegSc or OneRightOneInsideNegSc) are true, the test association is predicted as false , whereas, any other combination of features lead to a true association.

However, In case of neutral candidates, negation does not change the status of the association and it will remain false. Because of few numbers of neutral candidates in the produced corpus, considering the neutral candidates as negatives still leads to superior performance as will be seen in next section. As a result, if the status of the existence of neutral candidate was defined as

- "IsNeutralCand" A Boolean feature which is set as true when association candidate is neutral, while other situation is false.

The status of association can be calculated as below:

- $SNPTraitAssociation =$
  $(BothInsNegSc \vee OneLeftOneInsNegSc \vee OneRightOneInsideNegSc) \wedge$
  $\neg\ IsNeutralCand$

We compare the proposed negation neutral based algorithm (NNB) with the three kernel methods. The used kernel methods are global context kernel, local context kernel and sub-tree kernel. All of the three used kernel methods were trained with train part of the prepared corpus and were tested with test part.

In the next section, we will present the results obtained by the proposed method as well as those given by the kernel methods, so as a comparison can be made.

All of the kernel method experiments were carried out by a support vector with SMO [21] implementation. According to the experiments conducted via SMO approach and comparing the results to those of other implementations of SVM, e.g. libSVM, it was evident that SMO implementation was associated with a faster and better performance. Weka API was used as the implementation platform. A sample version of the proposed system is available online at the address (http://snpphenotypeext-nilg.rhcloud.com/).

### 3.3. Identifying level of confidence of SNP-phenotype association

There are genetic instructions for growing and developing all individuals, but environmental parameters also influence on the phenotype of a person through embryonic growth and life. Environmental parameters can be resulted by a range of effects including nutrition, weather, and disease and stress level. For example, the ability of tasting food is a phenotype, which is estimated as 85% affected via genetic inheritance [22]. On the other hand, this ability could be intervened by environmental parameters including dry mouth or lately eaten food.

The degree in which phenotype is determined by genotype is referred as "phenotypic plasticity" [23]. However, phonotypic plasticity is considered high if environmental factors have a strong influence. Conversely, if phenotypic plasticity is low if genotype can be used to reliably predict phenotype. Overall, the amount of influence that environmental factors have on a person's ultimate phenotype is a matter of serious scientific debate.

Different phenotypic plasticity as well as other effective unknown genetic components presents two explanations for why a GWA study reports on the importance of degree of confidence for these associations. Consequently, the linguist-based confidence of the reported association will have informative data leading to determination of phenotypic plasticity.

However, there is no available data source or automatic method for extracting level of confidence of the obtained results. Consequently, the presence of such a tool and data source is critical and can be applied to help researchers in reviewing the literature.

We have implemented a modality based supervised method (MMS) for identifying the level of confidence of the extracted association. The proposed method consists of a classifier initially was trained by the related modal markers, the mentioned p-value and the confidence level of the sentence which have been annotated in the corpus. And during the test phase initially modal markers and the container clause were identified. If the sentence doesn't have any modal markers or the entities were not located in the clause that contains the modals, the confident level will set to medium. Otherwise the level of confidence was determined by the trained classifier using the identified modal markers of the candidate sentence.

### 4. EVALUATION

In this section after presenting some statistical analysis regarding the number of different entities and linguistic-based negation cues and clause connectors in the corpus used for evaluation, cooperative validation results are demonstrated. We carried out two types of experiments, first the proposed method were carried out on train data set and were tested using test part and secondly 10 fold cross validation on the whole corpus. We have used three supervised kernel methods as benchmark. For this purpose, the support vector machine was used for this purpose.

The results revealed that the proposed method is superior to counterpart kernel methods. Besides, it eliminates the need for training data avoiding difficulties associated with this step done mostly by related experts.

- **Low proportion of complex sentences in the corpus.** The result of statistical analysis on clause connectors shows, 9.7% (=87/895) of the instances have concessive clauses. Furthermore, considering the table, it could be concluded that the most frequent connectors are "but" and "after".

Additionally, two third of the candidates have clause connector but this ratio is not significant in biomedical domain. While considering that biomedical scientific manuscript contain complex sentences usually mention different situation and condition. However, according to table 6, the average ratio of SNP and phenotype names per sentences is also weak.

- **Similar innate polarities of the sentences,** the polarity analysis shows that most of the sentences have innate positive polarity indicating an "association" between SNP and a phenotype. It is worth mentioning that the polarity analysis regarding associations was carried out on sentences without negation cue (Table 1). For instance the sentence beweak explains an "associated with" implication. Consequently, it has a positive innate polarity proving the existence of an association between operands:
  - In haplotype analysis, the haplotype combination of *rs2254298* A allele, *rs2228485* C allele and *rs237911* G allele was found to be significantly associated with an increased risk of **preterm birth** (OR=3.2 [CI 1.04-9.8], p=0.043).

### 4.1. Identifying the associations

In this section, the comparative results of the proposed method and local context kernel are presented in terms of F-score to calculate the positive classes.

**Table 1. Obtained comparative results for the proposed negation neutral based method (NNB) for the test corpus alongside to the obtained results for the three investigated kernel methods with non-neutral candidates (positive and negative-neutral class).**

| Method | LCK | Subtree kernel | NNB |
|---|---|---|---|
| F1 | 60.3% | 45.7% | 75.6% |
| Recall | 56.7% | 41.3% | 79.6 |
| Precision | 53.5% | 40.1% | 75.4 |

**Table 2. Obtained comparative 10 fold cross validation results for the proposed NNB method for the LCK kernel methods with two categories of candidates (positive and negative-neutral class).**

| Method | LCK | Subtree kernel | NNB |
|---|---|---|---|
| F1 | 94.2 | 91.5 | 97.4 |

The experiments were carried over two groups of the candidates. During the experiments whose results are shown in Tables 1 and 2, neutral candidates have been considered as a part of negative class of candidates as other relation extraction corpora.

To evaluate performance of our proposed method two other schemes are tested as well, namely, local context and sub-tree kernel methods. As it is shown in Table 1 and 2 the proposed method outperforms the mentioned schemes even when neutral class of samples is ignored. Moreover as tables show, local context kernel shows better performance in comparison with subtree kernel.

The role of neutral samples identification in improving performance of the NNB can be understood via examining Tables 1 and 2.
However, table 2 indicates f-measure values for all candidates in the corpus including positive, negative as well as neutral ones.

### 4.2. Forecasting level of confidence

In addition to the performed experiments for predicting the SNP-phenotype associations, a binary Bag of Word (BOW) method was performed over the corpus as a baseline method to predict degree of confidence for associations.

**Table 3. Obtained results for the calculating confident interval of the positive association of the test part of the SNPhenA corpus by Bag Of words and the proposed MMS method.**

| | Parameter | Low Level of confidence | Middle Level of confidence | High Level of confidence |
|---|---|---|---|---|
| **BOW** | F1 | 64.2% | 16.3% | 26% |
| | Recall | 64.6% | 14.5% | 27.7% |
| | Precision | 63.8% | 20% | 24.6% |
| **MMS** | F1 | 63.4% | 18.8% | 54.8% |
| | Recall | 51.9% | 10.9% | 64.7% |
| | Precision | 81.4% | 62.9% | 47.6% |

The achieved results are presented in Table 3. As the table shows, the best f-measure was achieved in those candidate expressions related to associations with a weak degree of confidence and the worst result was obtained in the medium degree of confidence. The reason for the weak result for the class with medium level of confidence is that there was small number of instances in the class. Moreover, better f-measure results of weak degree of confidence were determined there had been more trained instances. Moreover, the weak performance of the BOW method for two classes with stronger level of confidence can suggest that these classes overlapped with each other and that perhaps two classes of confidence would lead to better performance.

As table 3 shows, the proposed MMS method has better performance in comparison to the BOW method in terms of f-measure, precision and recall. In addition, as it is presented in the table, both methods have weak f-score, recall and precision in the category of middle level of confidences

**5.Discussion and Conclusion**

In this paper, we proposed a modality based SNP-phenotype association extraction method. The results demonstrate the superior performance of the proposed method. Additionally the results how the neutral candidates are important category of candidates that can be utilized for implementing better relation extraction methods. Moreover the achieved results show the importance of confidence level of the association as a linguistic-based factor can be used beside to existing methods to obtain more useful information. Although the proposed method shows promising results employing other feature can improve the performance of the confidence estimation of the extracted association. The estimated level of confidence of the association can be used beside to other factor such as abstract and paper confidence to define the overall confidence and credibility of the extracted association.

Although all existing relation extraction corpora and methods utilize **crisp** relations, the authors believe that it is not an efficient model for natural language's relations and they could be replaced with a better mathematical model called **fuzzy relations** (FR). Crisp relations deal with the binary relation between two entities in a sentence while FRs includes sets of fuzzy relations.

**BIBLIOGRAPHY**

[1] Gabor, T. Marth et al., "A general approach to single-nucleotide polymorphism discovery," *Nature genetics*, vol. 23, no. 4, pp. 452-456, 1999.

[2] Verspoor, K., Go Eun Heo, Keun Young Kang, and Min Song, "Establishing a baseline for literature mining human genetic variants and their relationships to disease cohorts," *BMC Medical Informatics and Decision Making*, vol. 16, no. 1, p. 37, 2016.

[3] Ashique, M., Tsung-Jung Wu, Mazumder, R.,& Vijay-Shanker, K., "DiMeX: A Text Mining System for Mutation-Disease Association Extraction," *PloS one*, vol. 11, no. 4, p. e0152725, 2016.

[4] Nature Education. [Online] (2016, July). HYPERLINK "http://www.nature.com/scitable/definition/phenotype-phenotypes-35" http://www.nature.com/scitable/definition/phenotype-phenotypes-35

[5] D Price, T., Qvarnstr, A.,& E Irwin, D.,"The role of phenotypic plasticity in driving genetic evolution," *Proceedings of the Royal Society of London B: Biological Sciences*, vol. 270, no. 1523, pp. 1433-1440, 2003.

[6] Eugene E. Loos, Susan Anderson, Day, Jr. Dwight H., Paul C. Jordan, and J. Douglas Wingate, *Glossary of linguistic terms*. Camp Wisdom Road Dallas: SIL International , 2004.

[7] Chapman, W., Bridewell, W., Hanbury, P. , Cooper, G.F., and Buchanan, B.G.,Evaluation of Negation Phrases in Narrative Clinical Reports, 2002.

[8] Joan L Bybee , Fleischman, S., *Modality in grammar and discourse*.: John Benjamins Publishing, 1995, vol. 32.

[9] McDonald, R., "Extracting relations from unstructured text," *Rapport technique, Department of Computer and Information Science-University of Pennsylvania*, 2005.

[10] Bokharaeian, B.,Diaz, A., Ballesteros, M., "Extracting Drug-Drug interaction from text using negation features", Sociedad Española para el Procesamiento del Lenguaje Natural, no. 51, pp. 49-56, 2013.

[11] Yifan Peng, C.O. Tudor, M. Torii, C.H. Wu, & K. Vijay-Shanker, "iSimp: A sentence simplification system for biomedicail text," in *Bioinformatics and Biomedicine (BIBM), 2012 IEEE International Conference on*, Oct 2012, pp. 1-6.

[12] Bokharaeian, B., Diaz, A., Neves, M.,& Francisco, V., "Exploring Negation Annotations in the DrugDDI Corpus," in *Proceedings of the Fourth Workshop on Building and Evaluating Resources for Health and Biomedical Text Processing*, 2014, pp. 84-91.

[13] Lee, k., et al., "BRONCO: Biomedical entity Relation ONcology COrpus for extracting gene-variant-disease-drug relations," *Database*, vol. 2016, p. baw043, 2016.

[14] Verspoor, k., et al., "Annotating the biomedical literature for the human variome," *Database*, vol. 2013, p. bat019, 2013.

[15] Horn, F.,Lau, AL.,& Cohen, FE, "Automated extraction of mutation data from the literature: application of MuteXt to G protein-coupled receptors and nuclear hormone receptors.," *Bioinformatics*, vol. 20, no. 4, Mar 2004.

[16] Ravikumar, K., Liu, H., D Cohn, J., E Wall, M., & Verspoor, K.,"Literature mining of protein-residue associations with graph rules learned through distant supervision," *Journal of Biomedical Semantics*, vol. 3, October 2012.

[17] Faisal, Md., Chowdhury, M., Lavelli, A.,& Fondazione Bruno Kessler, "Exploiting the Scope of Negations and Heterogeneous Features for Relation Extraction: A Case Study for Drug-Drug

Interaction Extraction," in *HLT-NAACL13*, 2013, pp. 765-771.

[18] Pyysalo, S., Airola, A., Heimonen, J., Bjorne, J., & F.and Salakoski, T. Ginter, "Comparative analysis of five protein-protein interaction corpora," *BMC bioinformatics*, vol. 9, no. Suppl 3, p. S6, 2008.

[19] Kim, J-D., and Ohta, T., and Tateisi, Y. and Tsujii, J., "GENIA corpus—a semantically annotated corpus for bio-textmining," *Bioinformatics*, vol. 19, pp. i180--i182, 2003.

[20] Chek Kim, L, and Miin-Hwa Lim, J.."Hedging in Academic Writing - A Pedagogically-Motivated Qualitative Study ," *Procedia - Social and Behavioral Sciences* , vol. 197, pp. 600-607, 2015, 7th World Conference on Educational Sciences. [Online]. HYPERLINK "http://www.sciencedirect.com/science/article/pii/S1877042815042019" [http://www.sciencedirect.com/science/article/pii/S1877042815042019](http://www.sciencedirect.com/science/article/pii/S1877042815042019)

[21] Thorsten, J., "Making large scale SVM learning practical," Universitat Dortmund, Tech. rep. 1999.

[22] Bokharaeian, B.,Diaz, A., & Chitsaz, H., "The SNPPhenA Corpus: An annotated research abstract corpus for extracting ranked association of single-nucleotide polymorphisms and phenotypes," in *second conference of signal processing and inteligent systems*, Tehran, accepted, 2016.

[23] Wooding, S.,et al., "Natural selection and molecular evolution in PTC, a bitter-taste receptor gene," *The American Journal of Human Genetics*, vol. 74, no. 4, pp. 637-646, 2004.